\documentclass[review]{fcs}

\usepackage{makecell}
\usepackage{float}
\usepackage{subfigure}
\usepackage{graphics}
\usepackage{marvosym}
\usepackage{url}

\title{Large Sequence Models for Sequential Decision-Making: A Survey}
\author{Muning Wen$^{*,1,2}$}
\author{Runji Lin$^{*,3,4}$}
\author{Hanjing Wang$^{1,2}$}
\author{Yaodong Yang$^{5}$}
\author{Ying Wen$^{1}$}
\author{Luo Mai$^{6}$}
\author{Jun Wang$^{2,7}$}
\author{Haifeng Zhang$^{3,4}$}
\author*{Weinan Zhang$^{1}$}
\address{$^1$Shanghai Jiao Tong University, $^2$Digital Brain Lab, $^3$Institute of Automation, Chinese Academy of Sciences, $^4$University of Chinese Academy of Sciences, $^5$Peking University, $^6$the University of Edinburgh, $^7$University College London}
\fcssetup{
  corr-email     = {wnzhang@sjtu.edu.cn},
}
\begin{abstract}
\begin{sloppypar}
Transformer architectures have facilitated the development of large-scale and general-purpose sequence models for prediction tasks in natural language processing and computer vision, e.g., GPT-3 and Swin Transformer.
Although originally designed for prediction problems, it is natural to inquire about their suitability for sequential decision-making and reinforcement learning problems, which are typically beset by long-standing issues involving sample efficiency, credit assignment, and partial observability.
In recent years, sequence models, especially the Transformer, have attracted increasing interest in the RL communities, spawning numerous approaches with notable effectiveness and generalizability.
This survey presents a comprehensive overview of recent works aimed at solving sequential decision-making tasks with sequence models such as the Transformer, by discussing the connection between sequential decision-making and sequence modeling, and categorizing them based on the way they utilize the Transformer.
Moreover, this paper puts forth various potential avenues for future research intending to improve the effectiveness of large sequence models for sequential decision-making, encompassing theoretical foundations, network architectures, algorithms, and efficient training systems.
As this article has been accepted by the Frontiers of Computer Science, here is an early version, and the most up-to-date version can be found at \url{https://journal.hep.com.cn/fcs/EN/10.1007/s11704-023-2689-5}.
\\
\end{sloppypar}
\end{abstract}


\begin{document}

\begin{sloppypar}

\clearpage

\section{Introduction}
Large sequence models, which feature a significant volume of parameters and auto-regressive data processing, have recently been instrumental in prediction tasks and (self-)supervised learning~\cite{liu2021self} in natural language processing (NLP)~\cite{nlp} and computer vision (CV)~\cite{computer-vision}, such as ChatGPT~\cite{qin2023chatgpt} and Swin Transformer~\cite{SwinTransformer}. 
Furthermore, these models, especially the Transformer~\cite{transformer}, have garnered substantial interest from the reinforcement learning community in the past two years, spawning numerous approaches as outlined in Section \ref{sec:how-transformer}.

In addition, large sequence models have emerged in the field of sequential decision-making and reinforcement learning (RL)~\cite{reinforcement-learning} with notable effectiveness and generalizability, as evidenced by Gato~\cite{reed2022generalist} and Video Pre-Training (VPT)~\cite{baker2022vpt}. 
These methods suggest the potential for constructing a large decision model for general purposes, that is, a large sequence model that can harness a vast number of parameters to perform hundreds or more sequential decision-making tasks, analogous to the way in which large sequence models have been leveraged for NLP and CV.

This survey focuses on most of the current works that leverage (large) sequence models, mainly the Transformer, for sequential decision-making tasks, while the application of various other types of foundation models in practical decision-making contexts could be found in the report by Sherry et al.~\cite{yang2023foundation}.
We offer an in-depth investigation of the role of sequence models in sequential decision-making problems, discussing their significance and how sequence models like the Transformer are related to solving such problems.
While surveying how current works utilize sequence models to facilitate sequential decision-making, we also analyze major bottlenecks toward large decision models currently with regard to model size, data and computation, and explore potential avenues for future research in algorithms and training systems to improve performance.

In the rest of this survey, Section \ref{sec:formulation} presents the formulation of prediction and sequential decision-making problems. 
Section \ref{sec:drl-for-dm} introduces deep reinforcement learning (DRL) as a classical solution for sequential decision-making tasks and examines three long-lasting challenges in DRL: sample efficiency problem, credit assignment problem, and partial observability problem.
Section \ref{sec:sm-for-dm} establishes the connection between sequence models and sequential decision-making, highlighting the promotion of sequence modeling regarding the three challenges raised in Section \ref{sec:drl-for-dm}.
Section \ref{sec:how-transformer} surveys most of the current works that leverage the Transformer architecture for sequential decision-making tasks and discusses how the Transformer enhances sequential decision-making in different settings as well as the potential for building large decision models.
Section \ref{sec:system} discusses the current progress and potential challenges regarding the system support for training large decision models.
Section \ref{sec:discussion} discusses current challenges and potential research directions from the perspectives of theoretical foundation, model architectures, algorithms, and training systems.
Finally, Section \ref{sec:summary} takes conclusions of this survey with the hope for more investigation into the emerging topic of large decision models.

\section{Formulation}\label{sec:formulation}
\subsection{Prediction Tasks}
Prediction in deep learning refers to the output of a neural network after it has been trained on a historical dataset and applied to new data when forecasting the likelihood of a particular outcome, e.g., image classification in CV and translation in NLP. 
For a classification task in CV, given an image $x$, the goal is to learn the estimation of the distributions $P(y{\mid}x)$, where $y$ is a potential label of $x$.
It is normally solved with discriminative models like Multi-layer Perceptron (MLP) or Convolution Neural Networks (CNNs)~\cite{kruse2022multi, lecun1989backpropagation, sarker2021deep}, extracting the high-dimensional representation $c(x)$ of the input image with convolution layers and estimating the distribution $P[y{\mid}c(x)]$.
For a translation task in NLP, an input sentence $\mathbf{x}$ is decomposed into a sequence with $n$ words $\left\{x_1,\dots,x_n\right\}$ to predict an output sentence $\mathbf{y}=\left\{y_1,\dots,y_n\right\}$. 
And the estimated distribution becomes $P(\mathbf{y}{\mid}\mathbf{x}) = P(y_1, \dots, y_n {\mid}x_1, \dots, x_n)$.
Besides, other NLP tasks like text generation, predicting the next potential word with previous contents, need to estimate only the distribution of $P(y_n{\mid}x_1,\dots,x_n)$ instead of $P(\mathbf{y}{\mid}\mathbf{x})$. 
Both $P(\mathbf{y}{\mid}\mathbf{x})$ and $P(y_n{\mid}x_1,\dots,x_n)$ could be modeled with sequence models like Recurrent Neural Networks (RNNs)~\cite{goodfellow2016deep} and their variants~\cite{hochreiter1997long, GRU}, which use their hidden states $h_{n-1}=h(x_{n-1},h_{n-2})$ to retain previous content and estimate the distribution of $P(y_n{\mid}x_n, h_{n-1})$ recursively.

\subsection{The Transformer}
As the state-of-the-art sequence model, the Transformer was originally designed for NLP tasks with an encoder-decoder structure. The encoder maps a sequence of tokens to latent representations, and then the decoder generates a sequence of desired outputs in an auto-regressive manner.
Besides, the encoder and decoder could also be used alone as models like Bert~\cite{bert} and GPT-3~\cite{brown2020gpt3}, which leverage the encoder and decoder architectures, respectively.
One of the most essential components in Transformer is the scaled dot-product attention, which captures the interrelationships of input sequences. The attention function is written as 
\begin{align}
    \text{Attention}(\mathbf{Q},\mathbf{K},\mathbf{V})=\text{softmax}\big(\frac{\mathbf{Q}\mathbf{K}^{T}}{\sqrt{d_{k}}}\big)\mathbf{V},
\end{align}
where $\mathbf{Q}, \mathbf{K}, \mathbf{V}$ correspond to the vector of queries, keys and values, which can be learned during training, and $d_k$ represents the dimension of $\mathbf{Q}$ and $\mathbf{K}$. Self-attentions refer to cases when $\mathbf{Q}, \mathbf{K}, \mathbf{V}$ share the same set of inputs. 
With the help of the attention mechanism, the Transformer abandons the recursive process of RNNs and estimates the distribution of $P(\mathbf{y}{\mid}\mathbf{x})$ or $P(y_n{\mid}x_1,\dots,x_n)$ more directly, enjoying higher computation efficiency.
Moreover, although the Transformer is initially designed for NLP tasks, it has the potential to be applied to CV tasks as well. By splitting an image into fixed-size patches, embedding each of them, and feeding the resulting sequence of vectors to a Transformer encoder, recent works have demonstrated remarkable performance of the Transformer in image classification tasks~\cite{dosovitskiy2020image}.

\subsection{Sequential Decision-Making Tasks}
Unlike prediction, sequential decision-making in deep learning refers to the process by which a neural network, known as an agent, infers a sequence of actions that can be used to interact with an environment and maximize its utility.
In most cases, a sequential decision-making problem is represented as a Markov decision process (MDP), $\langle \mathcal{S},\mathcal{A},r,p,\gamma \rangle$, that satisfies the Markov property~\cite{reinforcement-learning}
\begin{align}
p(s_{t+1}{\mid}s_{t},a_{t})=p(s_{t+1}{\mid}s_{0},a_{0},\dots, s_{t},a_{t}).
\end{align}
This property states that the current state of the process completely captures all the relevant information about the system's history, and thus the future is independent of the past given the current state~\cite{reinforcement-learning}.
In MDPs, $\mathcal{S}$ is the state space of the environment and $\mathcal{A}$ is the action space of agents. 
$r_{t}=r(s_{t},a_{t})$ is the reward function quantifying the instant utility of an agent executing an action $a_t\in\mathcal{A}$ on a specific state $s_t\in\mathcal{S}$. 
$p=p(s_{t+1}{\mid}s_{t},a_{t})$ is the transition probability of performing action $a_{t}$ on state $s_{t}$ at timestep $t$ and then transiting to state $s_{t+1}$. 
$\gamma$ is the factor used to calculate discounted returns 
\begin{align}
G_{t}=\sum_{k=0}^{\infty}\gamma^{k}r_{t+k}
\end{align}
that starts from timestep $t$.
At each timestep $t$, an agent takes an action $a_{t}$ based on the environmental state $s_{t}$. 
After execution, it receives an instant reward $r(s_{t},a_{t})$ and observes a new state $s_{t+1}$, whose probability distribution is $p(s_{t+1}{\mid}s_{t},a_{t})$. 
Following this process infinitely long, the agent earns a discounted return of $G_{t}$.
While $r(s_{t},a_{t})$ is the measurement of the instant utility of agents, $\mathbb{E}[G_{t}{\mid}s_{t}]$ is the expected cumulative utility starting in $s_{t}$, which is the objective of agents learning to maximize in sequential decision-making tasks.

\begin{figure*}[ht]
    \centering
    \includegraphics[width=\textwidth]{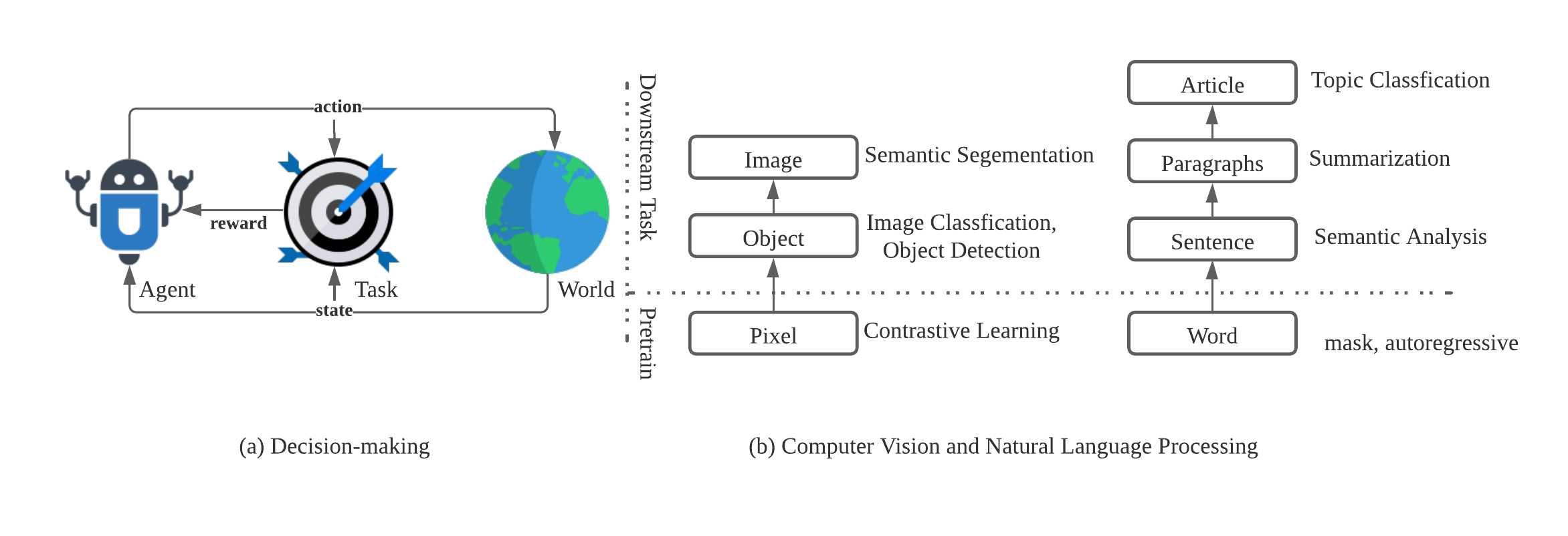}
    \caption{ The difference between sequential decision-making tasks and prediction tasks, such as CV and NLP. (a) A sequential decision-making task is a cycle of agent, task, and world, connected by interactions. (b) In prediction tasks, tasks form a hierarchical structure.}
    \label{fig:decision-representation-paradigm}
\end{figure*}

\section{Deep RL for Sequential Decision-Making}
\label{sec:drl-for-dm}
As a combination of deep neural networks and RL, deep reinforcement learning (DRL) has drawn much attention and emerged as a popular paradigm for solving sequential decision-making tasks~\cite{reinforcement-learning}. 
In recent years, its high potential has been demonstrated by a series of notable achievements, such as AlphaGo~\cite{alphago} and AlphaStar~\cite{alphastar}, which have beaten human experts at Go chess and StarCraft II.

In nearly all value-based RL methods, an agent measures the quality of an action under a specific state by learning an action-value function $Q_{\pi}(s_t, a_t)$,
\begin{gather}
    Q_{\pi}(s_{t},a_{t})=\mathbb{E}_{\pi}[G_{t}{\mid}s_{t},a_{t}].
\end{gather}
Specifically, the action-value function $Q_{\pi}(s_t, a_t)$ approximates the expected return starting from $s_t$ given that $a_t$ is selected, assuming the agent follows its policy $\pi$ thereafter.
A fundamental property of the value function is that it satisfies a recursive relationship between the expected return from the current state and the expected return from the following state, so-called the Bellman Equation~\cite{reinforcement-learning}:
\begin{gather}
Q_{\pi}(s_{t},a_{t})=r_{t}+{\gamma}{\max}_{a_{t+1}}Q_{\pi}(s_{t+1}, a_{t+1}).
\end{gather}
Through the utilization of the Bellman equation, Temporal-Difference (TD) methods~\cite{td-methods} can learn from incomplete sequences of experience by approximating the authentic value of the current state with the sum of the observed reward and the estimated value of the subsequent state.
Rather than waiting until the end of an episode as in Monte Carlo methods~\cite{williams1992simple}, TD methods thus update the value functions in a more efficient and incremental manner.
More specifically, in TD learning, an agent updates its $Q_{\pi}(s_t, a_t)$ by minimizing the mean square TD error~\cite{td-methods}:
\begin{gather}
\mathbb{E}_{\pi}[(r_{t}+{\gamma}{\max}_{a_{t+1}}Q_{\pi}(s_{t+1},a_{t+1})-Q_{\pi}(s_{t}, a_{t}))^2].
\end{gather}
In DRL, the $Q$ function could be approximated with neural networks and trained with gradient descent.
After learning an effective $Q$ network, agents' policies that maximize $\mathbb{E}[G_{t}{\mid}s_{t}]$ can be simply obtained by 
\begin{align}
\pi(s_{t})=\arg\max_{a_{t}}Q(s_{t},a_{t}),
\end{align}
which is widely adopted in many value-based methods like DQN~\cite{mnih2015human}.

While value-based methods learn to approximate the action values and then make decisions based on the estimates, policy-based methods, also known as policy gradient methods, learn the policy $\pi$ that selects actions directly without consulting a value function~\cite{reinforcement-learning}.
During training, policy gradient methods such as REINFORCE~\cite{williams1992simple} optimize the policy by maximizing the expected return below through gradient ascent.
\begin{align}
\mathbb{E}_{\pi}[\log{\pi(a_{t}{\mid}s_{t})}G_t]
\end{align}

Combining the value-based and policy-based methods, actor-critic methods~\cite{actor-critic} learn a state-value function $V_{\pi}(s_{t})$ as a critic to evaluate the quality of an actor given a state $s_{t}$, i.e., the expected return commencing from $s_{t}$ following the policy $\pi$:
\begin{align}
V_{\pi}(s_{t})=\mathbb{E}_{\pi}[G_{t}{\mid}s_{t}].
\end{align}
Similar to the $Q$ function, $V_{\pi}(s_{t})$ also satisfies the recursive relationship between the preceding and following states,
\begin{align}
V_{\pi}(s_{t})=\mathbb{E}_{\pi}[r_{t}+{\gamma}V_{\pi}(s_{t+1})],
\end{align}
and thus could be optimized by minimizing the mean square TD error as well:
\begin{align}
\mathbb{E}_{\pi}[(r_{t}+{\gamma}V_{\pi}(s_{t+1})-V_{\pi}(s_{t}))^2].
\end{align}
While updating the critic, the actor is optimized through policy gradient with the advantage function replacing the discounted return:
\begin{align}
\mathbb{E}_{\pi}[\log{\pi(a_{t}{\mid}s_{t})}A_{\pi}(s_{t}, a_{t})],
\end{align}
where the advantage function $A_{\pi}(s_{t}, a_{t})$ measures how well the selected action is compared with the actor's average performance.
\begin{align}
A_{\pi}(s_{t}, a_{t})
&=Q_{\pi}(s_{t}, a_{t})-V_{\pi}(s_{t})\nonumber \\
&=r_{t}+{\gamma}V_{\pi}(s_{t+1})-V_{\pi}(s_{t})
\end{align}

In model-based RL methods, a \textit{model} can be employed to predict how the environment will respond to agents' actions under a given state, by estimating the MDP's dynamics, $p(s_{t+1}, r_{t}{\mid}s_{t},a_{t})$~\cite{reinforcement-learning}.
The learning process for the model resembles a supervised learning task, but with data collected through real-time interaction with the environment.
Once the model is trained, it can be leveraged to generate action sequences via planning methods such as model predictive control (MPC)~\cite{camacho2013model}, or to generate imagined data as supplements to further enhance the value approximation or policy with direct RL, like what (deep) Dyna-Q does~\cite{reinforcement-learning, peng2018deep}.

However, despite DL having scaled RL to previously intractable problems, DRL is still not as widely applied in the real world as supervised or unsupervised learning. Several existing problems involving sample efficiency, credit assignment and partial observability have prompted extensive discussions~\cite{sample-efficiency,credit-assignment, DRQN}.

\subsection{Sample Efficiency Problem} 
\label{subsec:poor-sample-efficiency}
Poor data efficiency is one of the major restrictions of RL~\cite{sample-efficiency}.
In supervised learning, training data is labeled with ground truth $y$ so that models can learn to approximate the final distribution $P(y{\mid}x)$ of data from the beginning, which means models are fitting the same distribution during the training process.
Unlike supervised learning, conventional RL optimizes agents in a trial-and-error manner~\cite{reinforcement-learning}, which means the data distribution changes according to the current policy during the training process. Such a paradigm needs a series of loops to improve the quality of collected data and models alternately, i.e., data collection, model optimization, and data collection with optimized models.
For example, at the $k^{\text{th}}$ training epoch, agents collect dataset $\mathcal{D}_{k}$ with $\pi_{k}$, train a world model or value network with $\mathcal{D}_{k}$, update the policy and get $\pi_{k+1}$, which is used to collect $\mathcal{D}_{k+1}$ for the next epoch.
Since the policy $\pi$ for data collection is updated continuously, the collected dataset $\mathcal{D}$ is changing as well, which means the corresponding world model or value network is approximating a new data distribution in each training epoch and so is the policy.
Further, for environments with sparse rewards, the dilemma of poor sample efficiency will be more pronounced in the early training stages~\cite{mcfarlane2018survey, yang2021exploration}, since the initial random policies make it difficult to explore positive rewards and improve the quality of the dataset $\mathcal{D}$ and models.
Therefore, to guarantee the stability and effectiveness of the learning process, massive interactions with environments are indispensable in each epoch to explore enough positive rewards and fully reveal the new distribution.
However, these interactions can be expensive or even impossible due to safety concerns in real-world applications (e.g., autonomous driving~\cite{zhou2020smarts}, industrial scenario~\cite{qin2022neorl}).
Moreover, even slight differences between simulators and real environments (i.e., the reality gap) can lead to the vulnerability~\cite{reality-gap} of trained RL agents, constraining the current application of RL to a certain set of tasks.

\subsection{Credit Assignment Problem}
\label{subsec:credit-assignment-problems}
Mostly, the consequences of an action do not manifest immediately, requiring RL algorithms to capture the cause-and-effect relationship between a sequence of decisions and resulting rewards, known as the credit assignment problem~\cite{credit-assignment}, whose solution is crucial for effective and efficient algorithms.
While the simplest way to estimate the credit of a given state involves averaging its discounted sum of future rewards through Monte Carlo methods, such methods may suffer from high variance estimations and inefficient learning due to the randomness of trajectories~\cite{harutyunyan2019hindsight}.
To mitigate the variance, many RL approaches place more emphasis on TD methods with learned value approximation~\cite{td-methods}. 
But the approximation is likely to introduce bias, which spawns TD($\lambda$) methods to balance the bias-variance trade-off~\cite{td-methods}.
In most of the aforementioned methods, they rely solely on time as a metric of relevance: the more recent the decision, the more credit or blame it receives from a future result, which is heuristic in general and can hence be further improved by learning~\cite{reinforcement-learning, td-methods, harutyunyan2019hindsight}.

\subsection{Partial Observability Problem}
\label{subsec:partial-observable-problems}
In many real-world environments, it is common for parts of the state information to be unavailable and needed to be inferred by combining current observations with historical or other agents' observations~\cite{DRQN}.
This loss of state information can significantly confuse agents' decisions and hinder the development of effective decision-making agents.
For instance, in the case of an auto-driving car, providing only one image of a moment as the observation is insufficient to infer the speed of other vehicles, which is a crucial factor in deciding the next move.
Or in multi-agent settings, each agent's observations and experiences are often partial and potentially different from those of other agents, necessitating communication between agents to estimate the complete state of the system and make decisions.
This loss of full-state visibility expands the Markov decision processes (MDPs) to the partially observable Markov decision processes (POMDPs)~\cite{DRQN} for single-agent systems and decentralized partially observable Markov Decision Processes (Dec-POMDPs) for multi-agent systems~\cite{oliehoek2016concise}.
A common approach for addressing the partial observability problem is to model a sequence of observations with RNNs, expecting the missing information can be reconstructed during the training process~\cite{DRQN}.
However, information from early observations might be continuously diluted and even forgotten with the recursive function $h_{n}=h(x_{n}, h_{n-1})$ in RNNs, harming agents' performance when modeling long sequences~\cite{transformer}.

\section{Sequential Decision-Making as Sequence Modeling Problems}
\label{sec:sm-for-dm}
Fortunately, the challenges mentioned in Section~\ref{sec:drl-for-dm} could be addressed by treating sequential decision-making problems as sequence modeling problems and then be solved by sequence models.
In order to overcome these challenges, several researchers have attempted to simplify sequential decision-making tasks by transforming them into supervised learning problems, specifically, sequence modeling problems.
Imitation learning (IL), such as behavioral cloning (BC)~\cite{behavioral-cloning} and generative adversarial imitation learning (GAIL)~\cite{GAIL}, trains agents with the supervision of expert demonstrations, integrating advances in representation learning and transfer learning, e.g., the BC-Z~\cite{bc-z} or multi-modal interactive agent (MIA)~\cite{IL-Representation}.
However, the performance of IL depends heavily on high-quality expert data which is costly to obtain and conflicts with the increasing data requirements as the model size grows.
Upside-down reinforcement learning (UDRL)~\cite{UDRL} is a novel approach that transforms conventional reinforcement learning (RL) into a purely supervised learning paradigm. 
Compared with value-based RL, it reverses the roles of actions and returns during learning. 
Specifically, it employs undiscounted desired returns as network inputs, serving as commands to guide the agent's behavior. 
Thus, unlike conventional value-based RL, which learns a value model to evaluate the quality of each action and select the optimal one, UDRL learns to search for a sequence of actions that satisfy specific desired returns. 
By training the agent with pure SL on all past trajectories, UDRL circumvents the issues of sensitive discounted factors and the deadly trials arising from the combination of function approximation, bootstrapping, and off-policy training in traditional RL~\cite{UDRL,reinforcement-learning}.
Moreover, despite classical methods still being more effective in environments with perfect Markov properties, experimental results demonstrate that UDRL surprisingly exceeds conventional baselines, such as DQN and A2C, in non-Markovian environments~\cite{UDRL}. 
These results suggest that the general principles of UDRL are not restricted to Markovian environments only, indicating a promising direction for addressing sequential decision-making in a broader context.

\begin{figure*}[ht]
    \centering
    \includegraphics[width=0.75\textwidth]{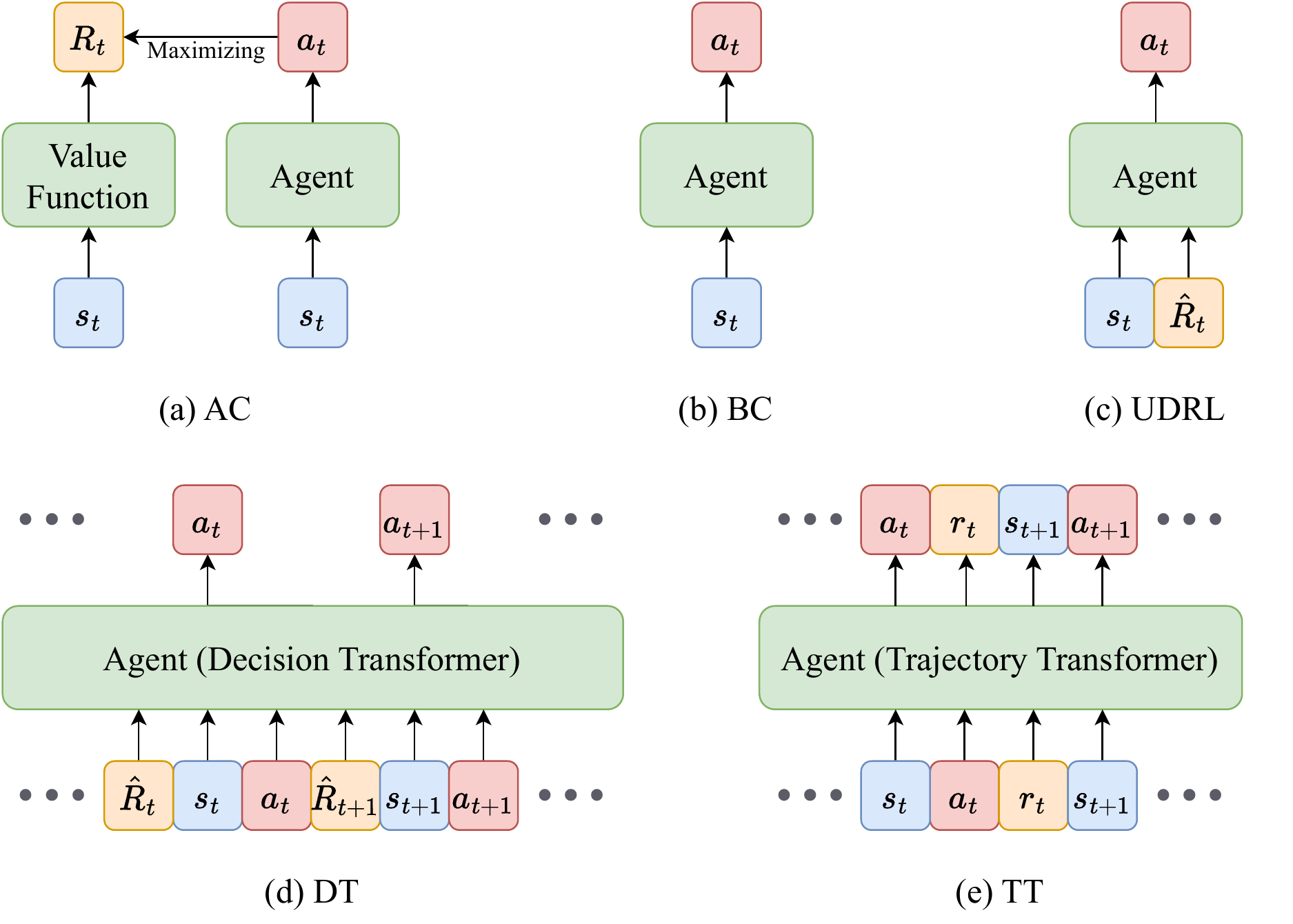}
    \caption{Paradigm comparison of conventional RL, IL, UDRL, DT and TT. (a) is a representative method of conventional RL, where $R_t$ indicates the estimated cumulative rewards with discount starting from $s_t$. (b) is a classic method in IL, i.e. Behavioral Cloning. In (c) and (d), $\hat{R}_t$ is the desired cumulative reward without discount. In (e), $r_t$ means the instant rewards after executing $a_t$.}
    \label{fig:method-comparison}
\end{figure*}

As a representative work, Decision Transformer (DT)~\cite{decision-transformer} frames RL problems as sequence modeling problems, which enables drawing upon the simplicity and scalability of the Transformer.
Based on the concept of UDRL, DT feeds a sequence of states, previous actions and desired returns to a GPT-like network and infers actions to achieve the desired returns, where the Transformer is served as a policy model.
Different from DT and UDRL, Trajectory Transformer (TT)~\cite{trajectory-transformer} maps transition sequences to shifted transition sequences entirely, incorporating states, actions and instant rewards, where the Transformer is served as a world model that captures the full dynamics of environments.
Although DT is a model-free method while TT is a model-based method, both approaches share a common foundation: treating each temporal trajectory as a continuous sequence of transitions and modeling it with the Transformer.
Based on this foundation, the Transformer could be used to infer future states, actions, and rewards, thus unifying many of the components that are typically required in IL, model-based RL, model-free RL, or goal-conditioned RL~\cite{trajectory-transformer}, e.g., predictive dynamics models in model-based methods, actor and critic in actor-critic (AC) algorithms~\cite{actor-critic}, and behavior policy approximation in IL.
Figure~\ref{fig:method-comparison} compares the paradigms between conventional RL, IL, UDRL, DT and TT.

\subsection{Improving Sample Efficiency}
As mentioned in Section \ref{subsec:poor-sample-efficiency}, conventional RL in a trial-and-error manner suffers from poor sample efficiency that limits its application in the real-world environment, because of the distribution shift at each epoch and inefficient exploration in the early training stages.
One of the directions to bypass this dilemma is pre-training~\cite{cang2021semi, wang2021lifelong, wang2022dirichlet}, leveraging previous experience from offline data to pre-train a suboptimal policy in a supervised manner and then fine-tuning it for downstream tasks, which appears in multiple recent works with the Transformer~\cite{cang2021semi,zheng2022online,meng2021offline}.
In this way, the trial-and-error process could start from the near-final stages with the suboptimal policy and leave aside most of the upfront exploration and interaction, reducing the online sampling epochs.
Especially for sparse reward settings, it can help skip the harrowing exploration in the early training stages.
With the strong generalizability of the Transformer architecture that has been validated in many practical results~\cite{brown2020gpt3,reed2022generalist,fan2022minedojo}, we can not only leverage the experience from the same task but also experience from many similar or related tasks, or even directly fine-tune the policies learned from other similar tasks~\cite{hu2021updet,zhou2021cooperative,wen2022multi,lee2022multi,xu2022prompting}.
From this perspective, the samples produced in different tasks could be stored for reuse when pre-training for new tasks, which improves the efficiency of sample utilization implicitly and largely.

\subsection{Effective Credit Assignment}
As discussed in Section \ref{subsec:credit-assignment-problems}, numerous conventional RL algorithms rely heavily on time as the primary metric for determining the cause-and-effect relationship between actions and rewards, which suffer from the bias-variance trade-off and require further improvement. 
Various works have explored better credit assignment through state association or learning additional reward functions to facilitate reward propagation over long horizons~\cite{ferret2019self, harutyunyan2019hindsight, mesnard2020counterfactual}.
In contrast, sequence models naturally embed this property in their architecture without requiring the explicit learning of extra reward functions~\cite{decision-transformer, trajectory-transformer, zheng2022online}. 
Furthermore, instead of assuming that recent actions receive more credit or blame, the attention mechanism can directly model the cause-and-effect relevance with undiscounted return sequences. 
Experiments conducted by Chen et al.~\cite{decision-transformer} have confirmed that this approach is more effective than conventional TD learning algorithms.

\subsection{Long Horizon for Partial Observability}
\label{subsec:long-horizon}
As described in Section~\ref{subsec:partial-observable-problems}, conventional RL methods have often relied on RNNs and their variants~\cite{hochreiter1997long,GRU} to recover the information lost from historical or other agents' observations~\cite{DRQN}. 
However, these methods still suffer from shortsightedness due to the limited capacity of hidden states, leading to the gradual dilution of early observations during recursion. 
For instance, given a sequence of observations $\left\{o_1,\dots,o_n\right\}$, an RNN models a policy as $\pi(\cdot{\mid}o_{n},h_{n-1})$, which prioritizes the last elements with limited capacity in the sequence and is difficult to build up long-term dependencies.
Compared with RNN-based methods, transformer-based methods model the policy as $\pi(\cdot{\mid}o_{1},\dots,o_{n})$, where the attention mechanism enables selective focus on specific parts of the sequence when making decisions~\cite{decision-transformer,meng2021offline, wen2022multi}. 
Specifically, the impact of an early observation, such as $o_{1}$, does not have to be diluted since the impact is determined by corresponding attention weights which are continually updated during training. 
Ablation experiments conducted by Muning et al.~\cite{wen2022multi} compare the performance of transformer-based and RNN-based policies and validate that the Transformer architectures enjoy a longer horizon than RNNs.

\section{How the Transformer Helps Sequential Decision-Making}
\label{sec:how-transformer}

The backbone architectures in machine learning have gone through several iterations, and the family of Transformers achieves a big convergence in the era of large-scale pre-trained models. 
From the linear model, Gaussian mixture model (GMM) and support vector machine (SVM) in the classical ML stage, to MLP, RNN and CNN in the DL stage, the Transformer and its variants rapidly become dominant for large-scale pre-training models in NLP~\cite{brown2020gpt3, bert}, CV~\cite{zhai2106scaling,SwinTransformer,SEER}, and multi-modal domains~\cite{CILP,DALL-E,UniversalTransformers, BEIT3}. 
Besides, with the appearance of a series of recent works listed in Table~\ref{tab:transformer-comparison}, the Transformer has also shown tremendous potential in the field of sequential decision-making.

\begin{table*}
\centering
\begin{center}
\begin{minipage}{\textwidth}
\caption{Detailed comparison between different Transformer-based methods for sequential decision-making.}\label{tab3}
\renewcommand{\arraystretch}{1.5}
\small
\begin{tabular*}{\textwidth}{@{\extracolsep{\fill}}lcccccc@{\extracolsep{\fill}}}
\toprule%
Method & Sequence & Prediction & Discretized Tokens & Benefit & Notes \\
\midrule
UPDeT~\cite{hu2021updet} & s & a & No  & \makecell[c]{Multi-task;\\Few-shot learning;\\Interpretability} & \makecell[c]{Model-free;\\Online;\\Multi-agent}\\
\midrule
PIT~\cite{zhou2021cooperative} & s & Q values & No  & \makecell[c]{Multi-task;\\Few-shot learning;\\Credit assignment} & \makecell[c]{Model-free;\\Online;\\Multi-agent}\\
\midrule
DT~\cite{decision-transformer} & rtg-s-a & a & No & \makecell[c]{Long sequence;\\POMDP;\\Credit assignment} & \makecell[c]{Model-free;\\Offline}\\
\midrule
TT~\cite{trajectory-transformer} & s-a-r(-rtg) & s-a-r & Yes & \makecell[c]{Long sequence;\\POMDP;\\Sparse-reward} & \makecell[c]{Model-based;\\Offline} \\
\midrule
GDT~\cite{furuta2021generalized} & $\psi(s,a)$-s-a & a & No & HIM problems & \makecell[c]{Model-free;\\Offline}\\
\midrule
PDT~\cite{cang2021semi} & s-a & a & No  & \makecell[c]{Few-shot learning} & \makecell[c]{Model-free;\\Pre-train}\\
\midrule
MADT~\cite{meng2021offline} & s-a & a & No & \makecell[c]{Multi-task;\\Long Sequence} & \makecell[c]{Model-free;\\Offline;\\Multi-agent}\\
\midrule
ODT~\cite{zheng2022online} & rtg-s-a & a & No & \makecell[c]{Few-shot learning} & \makecell[c]{Model-free;\\Online}\\
\midrule
MAT~\cite{wen2022multi} & s & a & No & \makecell[c]{Monotonic improvement;\\Multi-Task;\\Few-shot learning} & \makecell[c]{Model-free;\\Online;\\Multi-agent}\\
\midrule
MGDT~\cite{lee2022multi} & s-a-r-rtg & a-r-rtg & Yes & \makecell[c]{Multi-task;\\Few-shot learning} & \makecell[c]{Model-free;\\Offline}\\
\midrule
TrMRL~\cite{melo2022transformers} & s & a & No & \makecell[c]{Multi-task;\\Few-shot learning} & \makecell[c]{Model-free;\\Online;\\Meta-learning}\\
\midrule
PG-AR~\cite{fu2022revisiting} & s & a & No & \makecell[c]{Monotonic improvement} & \makecell[c]{Model-free;\\Online;\\Multi-agent}\\
\midrule
Prompt-DT~\cite{xu2022prompting} & rtg-s-a & a & No & \makecell[c]{Multi-task;\\Few-shot learning} & \makecell[c]{Model-free;\\Offline}\\
\midrule
BooT~\cite{wang2022bootstrapped} & s-a-r-rtg & s-a-r-rtg & Yes & Data Augmentation & \makecell[c]{Model-based;\\Offline}\\
\bottomrule
\label{tab:transformer-comparison}
\end{tabular*}
\end{minipage}
\end{center}
\end{table*}

In Section~\ref{sec:transformer}, we explore the empirical and theoretical advantages of the Transformer architecture as well as how it has become a popular choice in many state-of-the-art NLP or CV models. 
We then examine the development of the Transformer in the field of sequential decision-making, which can be divided into two parts. 
The first part focuses on recent works converting the reinforcement learning problem into sequential form to leverage sequence modeling for specific reinforcement learning settings, which will be surveyed in Section~\ref{sec:sequence-rl}. 
The second part concentrates on leveraging diverse data to pre-train a large-scale sequence model for various downstream sequential decision-making tasks, inspired by the tremendous success of NLP and CV, which will be discussed in Section~\ref{sec:pretrained-rl}. 
Finally, in Section~\ref{sec:big-decision-model}, we discuss the potential of building a large decision model and relevant characteristics that must be carefully considered.

\subsection{The Rise of the Transformer}\label{sec:transformer}
Transformers have a shown substantial impact on the progress of a large variety of machine learning tasks since the efficient expansion of model size helps harness massive amounts of data. Scale is a significant ingredient in achieving excellent results. Therefore, the model size is growing faster than ever before: benefiting from the Transformer architecture, large language models have scaled up from 340 million~\cite{bert} to 1.6 trillion parameters~\cite{SwitchTransformers} in a few years. As a result, the Transformers have outperformed previous standard networks (CNN and RNN) on numerous benchmarks and become general choices in the state-of-the-art model~\cite{BigTransfer,SwinTransformer}, e.g., image classification, semantic segmentation, text classification, text generation, question answering, image caption, etc.~\cite{BEIT3}.
Despite there being some attempts to build large pre-trained models based on CNN~\cite{BigTransfer}, Transformer architectures still take the dominant position in the field of large models. From empirical results and theoretical perspectives, Transformers have advantages in high parallelization~\cite{brown2020gpt3, bert}, scalability~\cite{zhai2106scaling, brown2020gpt3, kaplan2020scaling}, and appropriate inductive bias~\cite{Self-Attention}. In general, the advantages of Transformers supported by empirical evidence and theoretical analysis are summarized as follows. 

\textbf{Scaling law.}  The existence of the scaling law in Transformer architecture indicates that the loss scales as a power-law with model size, the amount of data, and the training computation~\cite{kaplan2020scaling, ghorbani2021scaling}.
There are several detailed and adequate experiments showing that the capacity of Transformer architectures increases smoothly following power law and the bigger models are more sample efficient in a series of tasks, such as  ViT-G in CV benchmarks~\cite{zhai2106scaling} and GPT in NLP benchmarks~\cite{kaplan2020scaling}. 
This finding has encouraged researchers to scale up their models to pursue higher performance.

\textbf{Higher throughput.} In the domain of sequence modeling, Transformer architectures exhibit superior throughput compared to RNNs, which possess inherent sequentiality: each hidden state is dependent on the previous hidden state.
This fundamental characteristic limits their ability to be parallelized across multiple GPUs, resulting in a considerable slowdown during training~\cite{goodfellow2016deep}. 
For instance, supposing a sequence with a length of $n$, a recurrent layer has to execute $n$ operations sequentially to backpropagate gradients for one training epoch. 
In contrast, the Transformer architecture offers more computational efficiency and parallelizability by avoiding sequential computation over time. 
Instead, it performs self-attention operations across the entire sequence at once, reducing the number of operations required for gradient backpropagation~\cite{transformer}.
This property helps Transformer-based methods to be trained at larger scale scenarios with acceptable computing budgets.

\textbf{Long-term interaction modeling ability.}  In terms of long sequence inputs, MLP suffers from the linear increase of the input layer dimension, vanilla CNN is limited by the local convolution kernel, and RNN is limited by an exponential decay of mutual information in the temporal distance~\cite{shen2019mutual}, which leads to difficulty in accurately modeling interactions between the long-spanning pairs. However, the attention mechanism enables the Transformer to efficiently handle very long sequences~\cite{brown2020gpt3}, which is discussed in Section~\ref{subsec:long-horizon} as well.

\textbf{More stable training process.} RNN frequently suffers from vanishing and exploding gradient problems~\cite{difficultyRNN}. On the contrary, Transformers are more robust in training. Researchers~\cite{kaplan2020scaling} observe the insensitivity of Transformers to some architectural hyper-parameters, which is vital for the training of large models considering the expensive training cost of conducting a hyper-parameter search.

\textbf{Efficient inductive bias.} Edelman et al.~\cite{Self-Attention} reveal that the inductive bias of self-attention is a creation of sparse variables to capture features of the input sequences. Olsson et al.~\cite{TransformerContext} demonstrate that the Transformer not only  memorizes  data patterns but also tries to conduct abstract reasoning.
Researchers also have provided theoretical analysis for the features of the Transformer, e.g., the inductive bias, sample complexity, and the generalization bound of the attention mechanism~\cite{wei2021statistically}. 
The others focus on measuring the model expressivity of the Transformer under the framework of universal function approximation and Turing completeness~\cite{UniversalTransformers,perez2018turing}.

\subsection{RL with the Transformer}\label{sec:sequence-rl}
Due to the noticeable effectiveness of DT and TT, many Transformer-based variants have recently emerged for sequential decision-making tasks, spanning from offline RL, model-based RL, meta RL, multi-agent RL, and goal-conditioned RL to agent architecture in the general RL setting. RL is suitable for the sequence modeling method, as a sequence of transition (trajectory) data includes information like environment states, actions decided by agents, and how the action affects the world, i.e., transition dynamics to the next stage, and task-specific rewards to measure the performance of behaviors. The major differences among these methods are listed in Table~\ref{tab:transformer-comparison}, such as the components in the sequence, how to process the sequence elements, benefits from sequence modeling, and specific reinforcement learning settings.

\subsubsection{Offline RL}
Offline reinforcement learning~\cite{levine2020offline} focuses on leveraging static datasets collected by behavior policy in various qualities without further interaction to train a better policy or evaluate it ~\cite{li2019perspective}. The sequence model provides a new perspective to tackle offline RL problems at the trajectory level. Because of the high similarity of the approach to using offline datasets with prediction tasks, this is the first area where sequence models are applied in RL. Decision Transformer (DT)~\cite{decision-transformer} adopts the reward condition from UDRL to boost the performance of the policy, and models a sequence of return-to-go, states, and actions. After supervised learning on offline data, DT demonstrates strong generalization to decode the better action when conditioned on an appropriately high return-to-go. However, it lacks a guiding principle to find an appropriately high return-to-go to achieve expert performance. To alleviate this issue, Multi-Game Decision Transformer (MGDT) introduces an expert classifier to conduct discriminator-guided generation for expert action.
Trajectory Transformer(TT)~\cite{trajectory-transformer} learns a world model to predict the future trajectory  from offline data and chooses the desired action by planning through beam search during execution. 
Extended from TT, Bootstrapped Transformer (BooT)~\cite{wang2022bootstrapped} boosts the sequence model training process with bootstrapping data argumentation. Although offline RL has advantages in data efficiency, sometimes an online fine-tuning process is necessary to achieve further performance improvements after offline learning.  However, DT is conservative due to the supervising manner, which impedes the exploration of the online process. For this purpose, the Online Decision Transformer (ODT)~\cite{zheng2022online} appends DT with hindsight return relabeling and entropy terms to encourage exploration. 

\subsubsection{Model-based RL}
Model-based RL~\cite{modelSurvey} utilizes  historical  data to build a world model to improve data efficiency and conduct safe planning. Sequence models use a historical sequence to predict the future and thus effectively reduce cumulative error. 
TransDreamer~\cite{transdreamer} and Dreamer with Transformers~\cite{dreaming} inherit the learning framework from Dreamer~\cite{hafner2020dream}, a notable MBRL algorithm, and simply change the backbone network architecture for agents and world models from RNN to Transformer. Benefiting from long-range modeling capability in Transformer, TransDreamer significantly surpasses Dreamer in benchmarks requiring complex memory.
Compared with learning a latent state representation and assuming that state distribution follows a prior distribution in TransDreamer, TT discretizes the continuous state and action into a sequence of discrete tokens, which represents a fixed width or quantile range of the original continuous space. Therefore, TT outputs an arbitrary probabilistic distribution of the next token conditioned on the historic discrete token sequence, significantly reducing the dynamic prediction error. 

\subsubsection{Meta RL}
Meta RL aims to train on diverse tasks to allow agents to adapt to new tasks quickly without much interaction in the environment.  
Pre-trained Decision Transformer (PDT)~\cite{cang2021semi} combines DT with semi-supervised learning to reduce the demand for labeled data through pre-training on massive unlabeled data, in which reward is regarded as the label in RL. Multi-Game Decision Transformer~\cite{lee2022multi} has no special design for meta RL pre-training, but simply uses a mixed dataset including several Atari trajectories with diverse performance. However, MGDT demonstrates rapid adaptation ability in out-of-distribution tasks with 1\% data to finetune. Prompting Decision Transformer (Prompt-DT)~\cite{xu2022prompting} leverages a prompting framework to enable rapid adaptation in offline RL, in which segments of task-specific demonstration are concatenated with input to guide agents to understand the new task.
TrMRL (the Transformer for Meta Reinforcement Learning)~\cite{melo2022transformers} employs a Transformer architecture to create an episodic memory to contextualize the policy, which is called the memory reinstatement mechanism. 
Generalized Decision Transformer (GDT)~\cite{furuta2021generalized} proposes a unified framework for hindsight information matching and a bi-directional DT which performs well in an offline one-shot imitation learning setting.

\subsubsection{Multi-Agent RL}
Multi-agent RL is proposed for the interactive scenario with several smart agents.  Sequence models in Multi-agent RL generally treat agents as a sequence, rather than a transition trajectory. Therefore, the interactions among agents can be captured by sequence modeling, which brings extra benefits, such as a monotonic improvement guarantee.  
Based on the DT, Multi-Agent Decision Transformer (MADT)~\cite{meng2021offline} extends it into multi-agent systems by directly applying the same architecture to independent agents with shared parameters.
While Wei et al.~\cite{fu2022revisiting} analyze the monotonic improvement property of auto-regressive policies in conventional multi-agent RL methods and propose the Auto-Regressive Policy Gradient (PG-AR) paradigm, Multi-Agent Transformer (MAT)~\cite{wen2022multi}, which is inspired by the Advantage Decomposition Theorem, incorporates the entire Transformer architecture and auto-regressive decision process into online multi-agent RL algorithms for monotonic improvement of joint policies and achieves state-of-the-art performance.

\subsubsection{Goal-conditioned RL}
Goal-condition RL~\cite{kaelbling1993learning, rudner2021outcome, liu2022goal} learns a general policy function to finish a series of simple tasks, for instance, to reach different goal states. TT can also be used in goal-conditioned RL by conditioning the goal state tokens in the planning process. Despite most of the sequence models in RL being GPT-style auto-regressive models, FlexiBiT~\cite{FlexFormer} uses a BERT-style bi-directional Transformer as backbone architecture to model the entire trajectory. Instead of predicting the next token from history, FlexiBiT is trained to predict some masked tokens given other tokens as context. Therefore, FlexiBiT is competent in goal-conditioned RL because it can predict the next action conditioned on the goal state by masking the intermediate sub-sequence. FlexiBiT provides a unified way to treat distinct RL tasks as different mask schemes, such as behavior cloning, offline RL, inverse dynamics, waypoint conditioning, goal-conditioning, etc. However, the current performance of the masked model is not satisfactory enough in general. Text-Conditioned Decision~\cite{Language-Conditioned} trains an agent to follow the instructions with the goal to take action.

\subsubsection{Agent Architecture}
Since the attention mechanism has some unique advantages, for instance, flexible input length and permutation invariance, the agents' backbone architecture based on Transformer enhances performance, which easily plugs into any conventional RL methods.
Universal Policy Decoupling Transformer (UPDeT)~\cite{hu2021updet} leverages the Transformer architecture to fit tasks with different observation and action configuration requirements.
Population Invariant agent with Transformer (PIT)~\cite{zhou2021cooperative} utilizes the Transformer architecture to achieve coordination transfer in universal scenarios.

\subsection{Scalable Pre-Trained Decision Models}\label{sec:pretrained-rl}
The huge amount of multi-modal interaction data on the Internet could be used to train a general model, helping agents understand their tasks and make various decisions according to humans' instructions in real-world applications~\cite{fan2022minedojo}.
While detailed comparisons between these Transformer-based sequence modeling methods are shown in Table~\ref{tab:large-model}.

\begin{table*}
\centering
\caption{We analyze what kind of data is used by these models (knowledge domain), how to understand the zero-shot generalization task (task indicator),  what kind of component the sequence model is deployed as (what to pre-train), how to pre-train the model, and how to use the pre-trained model. Below is an explanation of the abbreviations in the table: Language model (LM), language and vision model (LVM), and behavior cloning (BC). }\label{tab:large-model}
\renewcommand{\arraystretch}{1.5}
\small
\begin{tabular*}{\textwidth}{@{\extracolsep{\fill}}lccccc@{\extracolsep{\fill}}}
\toprule%
Methods & Knowledge Domain & \makecell[c]{Downstream\\Task Indicator}&  \makecell[c]{What to \\Pre-Train}& \makecell[c]{How to \\Pre-Train} & \makecell[c]{How to Use\\Pre-Trained Model}  \\
\midrule
Xland~\cite{team2021open} & Online tasks                  & Predicates       & Policy & RL & \makecell[c]{Zero-shot;\\Finetune}  \\
\midrule
MIA~\cite{IL-Representation}   & Offline human demo                & Text     & Policy & BC & \makecell[c]{Zero-shot;\\Finetune}   \\
\midrule
Gato~\cite{reed2022generalist} & \makecell[c]{Offline expert demo;\\ Multi-modal data}       & Prompt            & Policy & BC & \makecell[c]{Zero-shot;\\Finetune}  \\
\midrule
SayCan~\cite{saycan} & Pre-trained LM                 & Text & Perception & \makecell[c]{SL;\\RL} & zero-shot  \\
\midrule
Minedojo~\cite{fan2022minedojo}  & \makecell[c]{Internet video;\\Pre-trained LVM;\\} & Text  & Reward & SL & Online RL  \\
\midrule
VPT~\cite{baker2022vpt} &  \makecell[c]{Internet video;\\ Manual annotation}  & -           &  \makecell[c]{ Policy;\\ World Model} & BC & Finetune   \\
\midrule
LM-Nav~\cite{shah2022lmNav} & \makecell[c]{Pre-trained LVM;\\Pre-trained LM\\}                  & Text          & Perception & SL &  Search method  \\
\midrule
Inner Mono.~\cite{huang2022inner} & \makecell[c]{Pre-trained LM;\\Pre-trained VM}  & Text  & Perception & \makecell[c]{ SL;\\BC} & Zero-shot \\ 
\bottomrule
\end{tabular*}
\end{table*}

\subsubsection{Pre-Training for Sequential Decision-Making}

The essential differences between prediction and sequential decision-making problems make the current success of large sequence models in NLP or CV cannot be directly transferred to the latter. Because the sequential decision-making process involves a feedback loop, subtle changes in behavior would lead to severe data distribution shifts.
Therefore, new algorithms are demanded to learn stable representation, mitigate distribution shifts, and improve data efficiency.

We cannot expect that pre-training a single model would lead to strong generalization ability in all out-of-distribution tasks. 
Therefore, how to learn a universal and consistent representation for all the downstream tasks and minimize the distance between the training data distribution and the evaluation data distribution are the major issues that remain unsolved for effective large decision models with a reliable theoretical guarantee.

In general, representation learning and how to deal with distribution shifts are significant in pre-training, thus attracting interest from both CV and NLP.
Self-supervised learning contributes profoundly to the development of large models in representation learning.
NLP adopts the masking mechanism~\cite{bert} and auto-regressive process~\cite{brown2020gpt3}, while CV develops contrastive learning, such as SimCLR~\cite{simCLR} and MoCo~\cite{moco}. 
These methods help not only the utilization of unlabeled data but also produce a stable, informative, and consistent representation of data to speed up the downstream tasks.
Prompting~\cite{xu2022prompting}, a recently proposed training paradigm in NLP is challenging the typical pre-train and fine-tune paradigm.
Briefly, prompting methods transfer downstream tasks into some prompt templates.
In essence, prompts convert the evaluation distribution into the training distribution, therefore being a promising solution for the zero-shot setting.

However, this issue is more challenging in the sequential decision-making domain.
First, as Figure \ref{fig:decision-representation-paradigm} shows, the relationship between pre-trained tasks and downstream tasks is hierarchical in prediction problems, while it is cyclic in sequential decision-making problems.
This difference makes \textit{learning what kind of representation} and \textit{how to organize the downstream tasks} remains an open problem.
Second, since the decision made by the agent would affect the world, the sequential decision-making problems suffer from severe distribution shifts, which impede generalization~\cite{understanding}.  
This problem is abstracted as the auto-induced distributional shift~\cite{auto-induced-distributional-shift}, which means the output of a system causes a change in input data. 
Although researchers provided a theoretical analysis framework of the distribution shift from the difference between behavior policy and training policy~\cite{offline-shift}, we should consider more factors, such as the different world dynamics and task objectives in downstream tasks.

\subsubsection{Data Collection for Pre-Training}
Data, model size, and computing are the three main performance bottlenecks, according to empirical findings~\cite{kaplan2020scaling}.
There are two potential research topics for expanding the available data in the sequential decision-making domain, while model size and computation are discussed in Section \ref{sec:transformer} and \ref{sec:system}, respectively.

The first focuses on creating a procedural framework to generate a wide spectrum of tasks and scenarios in simulators to eliminate bottlenecks from a limited number of human-designed tasks~\cite{team2021open}.
Massively diverse and flexible tasks provide essential knowledge to develop skills in logical reasoning, understanding, planning, and memory to solve new complex sequential decision-making problems.
However, since all the training tasks are generated in the same format, how to eliminate the gap between training tasks and downstream tasks remains an open problem.
Proposing efficient algorithms to transfer the skills from simulation to the real world~\cite{kaspar2020sim2real} for large model settings or constructing realistic but scalable simulators to reflect the real world as much as possible~\cite{tancik2022blocknerf} are promising directions.

The second way focuses on leveraging diverse, large but static datasets without further interaction to train a sequential decision-making system, termed offline reinforcement learning~\cite{levine2020offline}.
This paradigm greatly extends the boundaries of applications, as interaction with the environment is infeasible, expensive, and unsafe in most real-world tasks, e.g., automatic driving, recommendation systems, and robotics.
Although offline algorithms have the aforementioned advantages, offline RL faces a series of challenges, including smoothly transitioning from offline pre-training to online fine-tuning to achieve better performance~\cite{nair2020awac,mao2022moore}, hyper-parameter sensitivity, and a lack of an efficient evaluation method to search for better hyper-parameters and examine policy robustness.

\subsubsection{Recent Advances in Scaling Pre-Trained Decision Models}
As shown in Table~\ref{tab:large-model}, there are several attempts in pre-trained decision models trying to answer the aforementioned questions. Researchers utilize diverse datasets from distinct knowledge domains, including online interactions with environments or offline demonstrations, to pre-train different components in decision systems. These methods are characterized by what kind of component in decision systems is pre-trained and how to use the pre-trained model.

\textbf{Pre-Training for Policy.} 
A policy with ideal initialization can mitigate the exploration problem better than learning from scratch, which is verified in many scenarios, e.g., Go and Starcraft. Therefore, pre-training a policy on enough diverse and massive experience data can improve data efficiency in the new downstream tasks. In an important attempt, Gato~\cite{reed2022generalist} pre-trains a single large model on multi-modal data to master hundreds of tasks, including sequential decision-making tasks, image captions, chitchat, etc. By simple imitation learning on expert demonstrations, Gato successfully pushes the model parameter scale in the sequential decision-making domain to the billion level. Also, it avoids some primary challenges in sequential decision-making, such as learning ability with suboptimal offline data and high data efficiency in the online fine-tuning process.
VPT pre-trains a single sequence model to imitate human player behavior from massive YouTube videos. The pre-trained model served as a general behavioral prior, showing zero-shot capabilities and making exploration easier and more efficient in fine-tuning.
MGDT focuses more on policy transferring or few-shot adaptation across multiple tasks with the strong generalizability of Transformer architectures. Crucially, Gato, VPT, and MGDT all show scaling law in the RL field, indicating the pursuit of large decision models is promising.

\textbf{Pre-Training for Reward Function.} The reward function plays a key role in sequential decision-making systems, and defines the target of tasks or the preferences over a series of different policies.
On account of the importance of the reward function, pre-training a reward function for downstream tasks helps RL work on completely new tasks without human design.
Minedojo~\cite{fan2022minedojo} pre-trains a large language and vision model to approximate reward functions and guide the online reinforcement learning to generalize into unseen task instructions.

\textbf{Pre-Training for World Model.} In a specific setting, a world model simulates the environment that agents interact with, which is a reusable component shared by a series of tasks. Although TT provides a promising tool to train a world in a sequential manner for robotics or other similar scenarios, to the best of our knowledge, there is no study to fill this gap. However, an inverse world model has been introduced to increase the diversity and quantity of offline data. The inverse world model in VPT~\cite{baker2022vpt} expands the sources of data from laboratories to large-scale realistic internet information produced by humans. Specifically, VPT collects a relatively small dataset with game videos played by volunteers labeled with action sequences and a large set of videos without action labels from YouTube. Then an inverse dynamic world model is trained on the small dataset to label massive YouTube videos, and human demonstrations with action labels are used to pre-train a policy.

\textbf{Pre-Trained Multi-Modal Perception Model for Sequential Decision-Making.} To attain agents with general skills, basic common sense is indispensable, such as the ability to recognize objects from pictures, understand semantics from text, and decompose a task into steps~\cite{zhou2022rehearsal}.
Multi-modal algorithms~\cite{shah2022lmNav, huang2022inner, saycan} improve data efficiency by transferring knowledge from off-the-shelf pre-trained large sequence models in the language or vision domains, rather than cultivating basic ability in a trial-and-error manner.
SayCan combines a value function and a pre-trained language model to control a real robot, following task instructions~\cite{saycan}. Specifically, the value function figures out what action can be completed, while the pre-trained language model figures out whether this action is appreciated to achieve the task goal.
Huang et al.~\cite{huang2022language} decompose a complex task into several simple goals, speeding up the learning process of downstream tasks and showing strong generalization.
Inner Monologue~\cite{huang2022inner} implements a closed-loop feedback control system for robotics based on a pre-trained language model and a collection of perception models, thinking of completing the entire task as a conversation. Perception models provide scene information. An agent driven by the language model decides what to do next and inquires for human feedback to give the correct response, and the human describes the tasks and interacts with agents. It is observed that all participants in a conversation give information in language form.
LM-Nav~\cite{shah2022lmNav} achieves impressive performance in open real-world robotic navigation tasks, powered by large pre-trained models of language, vision, and action. The language model is responsible for converting navigation commands into a series of landmarks, and the vision-and-language model grounds the landmarks in the topological map. The knowledge from the pre-trained model significantly eliminates the bottleneck caused by limited language-annotated robot data.

\subsection{The Next Step: Large Decision Models}\label{sec:big-decision-model}
Gato~\cite{reed2022generalist} and VPT~\cite{baker2022vpt} have shown the potential of building large decision models for general purposes in the field of sequential decision-making, like what large sequence models have done for NLP and CV tasks.
However, to build a large decision model, some modifications in architecture are significant with increasing data and model size, while naively scaling up models might fail as the number of parameters increases. 
That is, with the same volume of data and parameters, the network architecture can be the determining factor to improve the performance of large decision models. 
In this section, some important characteristics are listed since they can serve as consultative principles when designing network architecture for large decision models in the future.
Noticed that the Transformer and its variants are suggested to be promising candidates recently, but any other model architectures~\cite{bai2018empirical,MLP-mixer,Perceiver_io} meeting the requirements below are still worth an exploration.

\subsubsection{Multi-Task} 
To take full advantage of high-capacity models, how to utilize data from diverse tasks is critical for generalization. 
Some techniques in model architecture have been investigated, e.g., transfer learning can be accomplished with the mixture of experts (MoE)~\cite{MoE} and modularization~\cite{soft-modularization,Pathnet}.
Related research can help large models in the sequential decision-making domain attain better general intelligence.

\subsubsection{Sparse Activation} 
When decision models are scaled to extreme sizes, a computing request involving the full set of parameters can be incredibly expensive and inefficient.
However, for dense models, each piece of inference or training data activates the entire set of model parameters, resulting in high training and inference costs and latency.
Therefore, the parse activation~\cite{SwinTransformer,lepikhin2020gshard} methods are proposed to balance model size and performance. 
For each piece of data, only a subset of the parameters is activated to process the input in a sparse model during training and testing. 
Even if sparse models usually have more parameters when compared to their equal-quality dense alternatives~\cite{rajbhandari2022deepspeed}, their training or inference cost and latency are significantly reduced.

\subsubsection{Multi-Modality} 
Data in different modalities provides information from distinct perspectives.
Model architectures supporting multi-modality are capable of a broader range of applications and more complicated interactions.
Vision endows agents with the ability to observe, make a reasonable response~\cite{team2021open}, and form general knowledge about the shape of objects~\cite{shah2022lmNav,baker2022vpt}.  
Natural language instructs agents and deepens their understanding of the new tasks~\cite{huang2022language}, divides the high-level goal into detailed steps~\cite{shah2022lmNav,fan2022minedojo, huang2022inner} and provides a natural interface for agents to cooperate with humans~\cite{IL-Representation}.

\section{Training Systems}\label{sec:system}
In this section, we discuss the systems that can support the training of large decision models based on sequence models.
Sequence models, especially with Transformer architectures, have achieved substantial improvements in accuracy and generalizability by scaling up, usually following scaling laws~\cite{kaplan2020scaling,zhai2106scaling,ghorbani2021scaling}.

{\bf Model and Data Scaling}. Although based on highly different test suits and tasks (image classification~\cite{zhai2106scaling}, language translation~\cite{ghorbani2021scaling}), prior research reports that Transformers exhibit highly predictable scaling patterns in many of these tasks. As proposed, the test loss of the model when saturating (given enough training data) follows a general form:
\begin{equation}
    \label{eq:pwr-law}
    \hat{\mathcal{L}}(N_c) = \alpha (N_c/N)^{-\beta} + \mathcal{L}_{\infty},
\end{equation}
where ${\alpha, \beta, L_{\infty}}$ are fitted parameters depending on tasks and data. $N_c$ is the number of non-embedding parameters. $N$ is a fixed normalization term for $N_c$ and $L_{\infty}$ represents the irreducible part of the loss in scaling due to data noise. Besides the model saturating law revealed by Equa. \ref{eq:pwr-law}, empirical results~\cite{kaplan2020scaling} show that models have better data efficiency and can benefit from a larger dataset when scaling up. In the language translation tasks~\cite{kaplan2020scaling}, the number of training data that saturates models when scaling can be fitted by a sub-linear power-law (dataset volume $D\sim N^{0.74}$). To get the most out of scaling, the sizes of the model and data are required to be expanded simultaneously, imposing new challenges to the design of efficient training systems due to issues such as massive memory and computational budget demanded.

{\bf Scaling decision models}. While there exists a fruitful line of work on scaling behaviors of vision and language Transformers, that of large decision-making Transformers is still under-explored. Even though both are trained in a supervised manner, recent researches~\cite{IL-Representation,reed2022generalist} on large decision models report different scaling patterns, not to mention their reinforced counterparts, which often have a stronger dependency on resource-demanding online data generation. Efficient training of large decision models is not a trivial problem, and the lack of a handy training toolkit and systems for large decision models is holding back more research forces from entering this area.

\subsection{Existing Challenges}

\subsubsection{Hybrid Parallelism}
\begin{figure*}[ht]
    \centering
    \includegraphics[width=\textwidth]{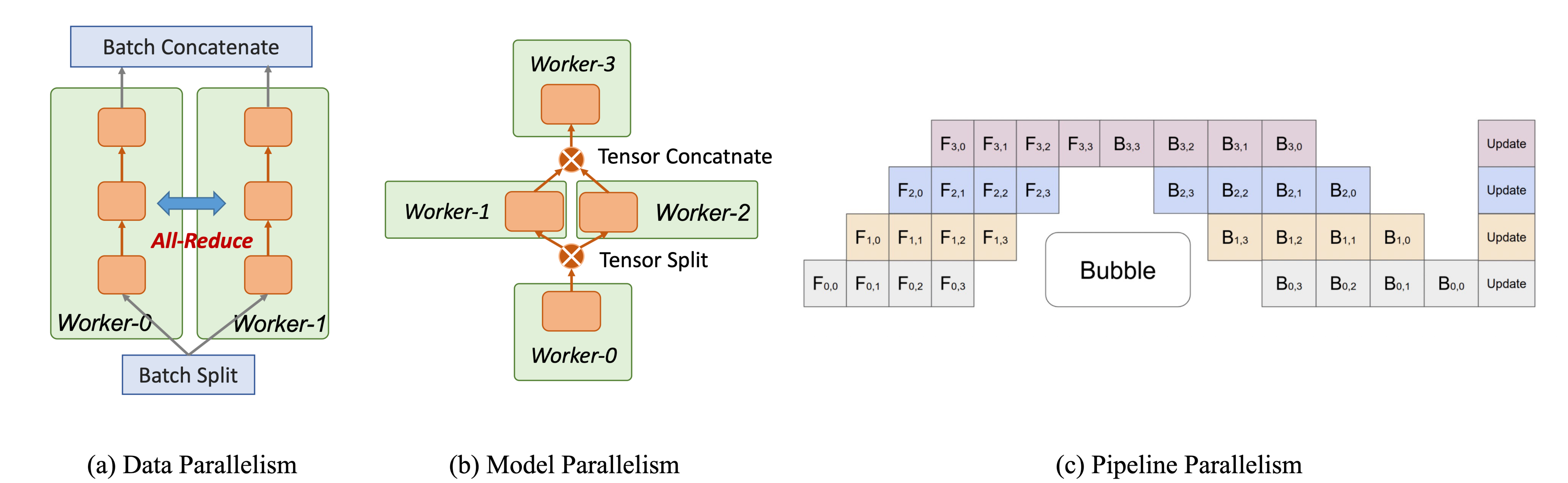}
    \caption{(a) shows a data-paralleled three-layer model with a parallel size of 2. Data Parallelism (DP) creates replicas of the entire model across the cluster, with each device holding one (or more) of these replicas. (b) illustrates the same three-layer model being assigned to 4 physical devices under Model Parallelism (MP), with a layer-wise (vertical) slicing schema and a horizontal slicing scheme on the second layer (the $2$-nd layer being internally sliced and assigned to worker-$1$ and worker-$2$). MP splits the model either horizontally (inside a layer, where Tensor Parallelism is often involved since parameters like weights are sliced, e.g., split matrix multiplication into operations into sub-matrices) or vertically (layer-level slice). (c) GPipe~\cite{huang2019gpipe}: A 4-layer model assigned to 4 physical devices (the vertical axis) with a parallel parallelism schema. Parallel Parallelism (PP) combines DP and MP by slicing the model vertically into chunks, mapping them to different devices, and splitting the mini-batch input into micro-batches fed into the pipeline sequentially to reduce bubbles (device under-utilized periods). {\bf Hybrid Parallelism}: Though PP has already been a hybrid of DP and MP, it can be further integrated with DP inside a parallel schema by serving multiple homogeneous pipelines (parameters can differ depending on the synchronization schema), orchestrated as a hybrid parallelism schema. A hybrid parallelism schema is often a combination of DP, MP and PP to have fine-grained placement and execution plans based on diverse IO, memory, and computing characteristics of different parallelism methods with an overall optimization goal of efficiency.}
    \label{fig:paralliially to reduce bulsm}
\end{figure*}

Gigantic sequence models can contain trillions of parameters. 
These parameters consume tremendous  amounts of memory and must be distributed to multiple devices. 
The distributed execution is usually achieved through a hybrid parallelism scheme that combines data parallelism, model parallelism, and pipeline parallelism, as shown in Fig.~\ref{fig:paralliially to reduce bulsm}.
To train large sequence models, training systems must have effective ways to optimize hybrid parallelism schemes and distribute computation to multiple devices.

\subsubsection{Large Datasets and Massive Environments} 
\begin{figure}[ht]
    \centering
    \includegraphics[width=0.95\linewidth]{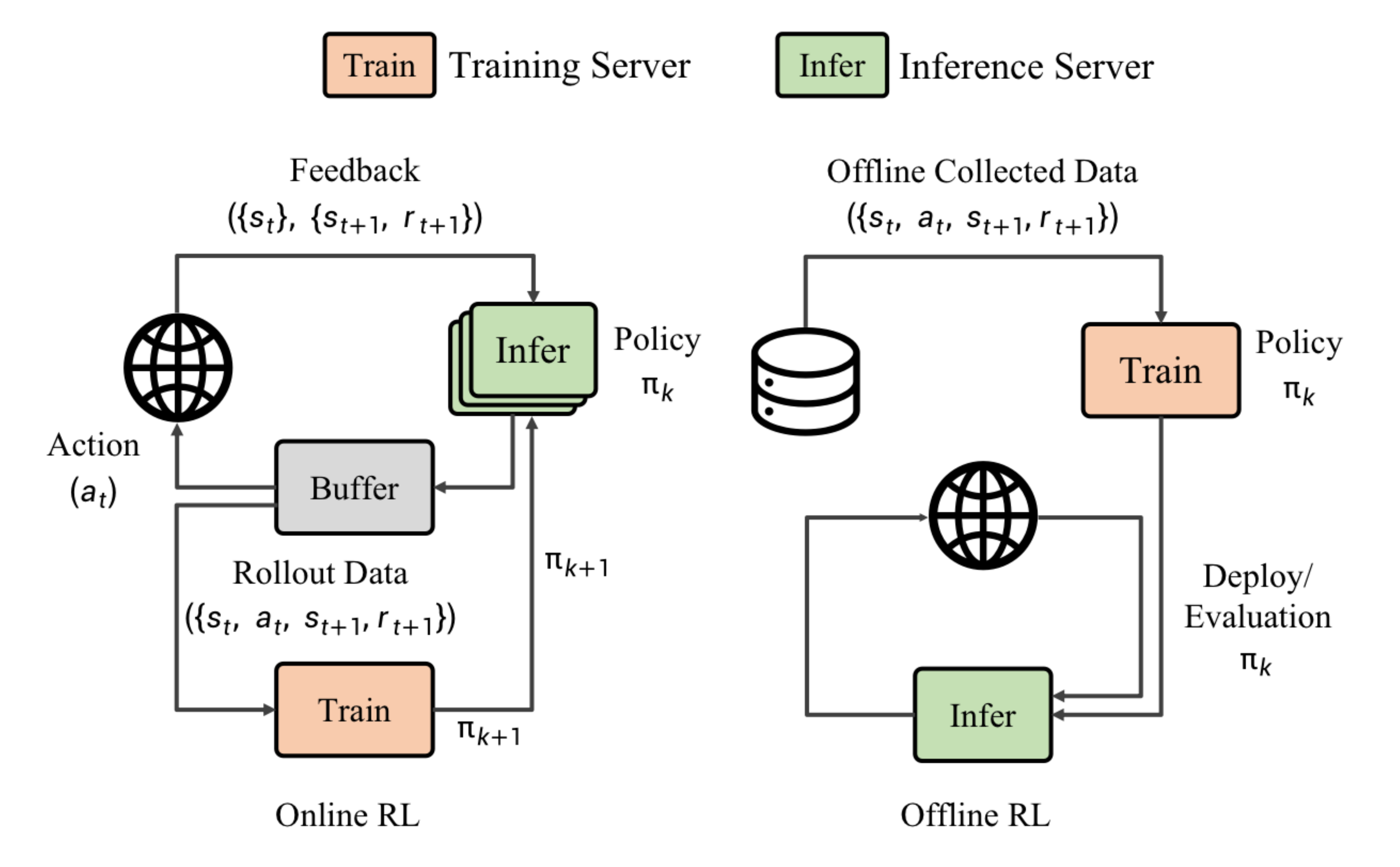}
    \caption{The data-flow comparison between the paradigms of offline RL and online RL, where offline pre-training relies on large datasets and online fine-tuning requires parallelizing massive environments to accelerate online interaction and data collection. Moreover, the online fine-tuning phase imposes more communication pressure due to strict parameter synchronization requirements between inference and training servers. }
    \label{fig:paradigm-diff}
    \vspace{-10pt}
\end{figure}

Sequence models need to be pre-trained using offline datasets and fine-tuned (i.e., few-shot learning) for downstream tasks by interacting with environment simulators online, as shown in Fig.~\ref{fig:paradigm-diff}.
The pre-training datasets can be as large as 500 billion tokens~\cite{brown2020gpt3}.
The parallel execution of environment simulators is also challenging.
These simulators need to be parallelized using thousands of CPUs (and even GPUs), thus producing a sufficient workload for fine-tuning sequence models.

\subsection{Hybrid Parallelism Systems}
For training large Transformer architectures, many hybrid parallelism systems have been proposed.
For example, GPipe~\cite{huang2019gpipe} introduces pipeline parallelism for the Transformer, DeepSpeed partitions the states of the optimizer to reduce communication overhead, and Colossal-AI~\cite{bian2021colossal} attempts to automatically parallelize the training of gigantic neural networks.
Though promising, these systems fail to fully support large decision models for several reasons.

\subsubsection{Lack of Designs for Synchronizing Model Checkpoints}
Existing hybrid parallelism systems are often designed for offline training scenarios where models are trained for a long time, and model checkpoints are deployed into inference servers only once.
In the scenarios of training decision models, the models need to be continuously checkpointed and repetitively deployed to the inference servers (i.e., model synchronization); otherwise, the inference servers will have stale models that offer sub-optimal performance when interacting with environment simulators.
As large decision models can have trillions of parameters, continuously checkpointing and deploying large model checkpoints can incur severe network bottlenecks, making the training of large decision models prohibitively expensive.

\subsubsection{Lack of Designs for Handling Simulation Environments}
Existing hybrid parallelism systems are originally designed for processing training datasets stored on disks, and they are ill-suited to processing simulation environments that continuously produce training samples in memory.
Different from on-disk training datasets, simulation environments can return highly complex nested observations, and those observations are produced at dynamic rates (i.e., observations are produced at a different rate at the beginning of training because initial decision models often offer insufficient performance).
Integrating these environments into existing offline-oriented hybrid parallelism systems requires non-trivial research and implementation efforts.

\subsection{Distributed RL Systems}
Distinct from building hybrid parallelism systems, practitioners have also made parallel efforts in designing distributed RL systems.
Ray allows multiple RL tasks (e.g., simulating environments or training RL models) to be dynamically dispatched to CPUs and GPUs. Impala~\cite{IMPALA} adopts an Actor-Learner architecture where actors (consisting of an inference model and an environment simulator) produce trajectories in parallel, and learners replicate RL models on multiple GPUs. Seed-RL~\cite{espeholt2019seed} further speeds up actors by allowing GPUs to be effectively used in model inference.

However, there are non-trivial challenges for existing variants of distributed RL systems to accommodate the training tasks of large decision models.
We observe several reasons for this problem.


\subsubsection{Lack of End-to-End Performance \\ Optimization}
Training a large decision model requires a complex pipeline. Specifically, the model requires (1) processing large training datasets first, (2) interaction with environments, and (3) using hybrid parallelism to partition large model states finally.
This pipeline requires various techniques to optimize hardware performance: there are techniques for using GPUs to speed up dataset processing and environment simulation or parallelize the computation of large tensors.
These days, all these techniques are applied in an isolated manner, and they are not coordinated in existing RL systems.
Such systems thus lack end-to-end performance optimization, leaving underlying hardware resources inefficiently utilized.

\subsubsection{Lack of Automatic Resource Management}
An enabling scenario for large decision models is multi-task pre-training.
These tasks need to be driven by different environments, and the pre-trained models can adopt a mixture-of-expert architecture.
These days, users must \emph{manually} allocate GPU resources to different environments, and further reserve GPUs for pre-trained models.
This manual resource allocation is, however, tedious and often sub-optimal, and we anticipate future distributed RL systems to realize fully automatic resource management, making them capable of supporting large-scale multi-task pre-training.

\section{Discussion and Future Prospects}
\label{sec:discussion}
\subsection{Theoretical Foundation}
Although converting RL problems into sequence modeling problems has yielded numerous benefits recently, it has also resulted in a loss of theoretical guarantees for policy optimization, in contrast to traditional RL approaches. 
While satisfactory performance has been achieved in some experiments, this superiority is heavily dependent on the generalization ability of network architectures, data quality, and specific problem scenarios. 
For instance, DT-type algorithms might experience significant degradation in environments with high randomness destabilizing the desired returns. 
The lack of effective theoretical analysis and guarantees for policy optimization constrains further improvement of decision models. 
Therefore, it is highly meaningful to research the organic integration of sequence modeling methods with traditional RL methods that offer theoretical guarantees in the future.

\subsection{Network Architectures}
In terms of network architecture, most of the RL methods directly rely on vanilla Transformers from NLP and CV without customized design, leaving ample room for performance improvement. 
Developing customized Transformer architectures primarily involves defining sequences, designing tokens, targeted attention calculation, and employing MoE layers. 
Consequently, promising research directions include integrating RL-specific semantics into the token design, combining attention masks with the Markov properties, and allocating MoE specifically to sequential decision-making tasks. 
Additionally, in-context learning is an important feature of large language models that often require lengthy sequences to emerge. 
Leveraging the Markov properties of sequential decisions to reduce computation complexity from quadratic to linear is a highly valuable research problem with the potential to facilitate in-context learning. 
Lastly, the recent surge of diffusion models yields novel implications for modeling decision sequences.

\subsection{Algorithms}
Despite recent advancements, many RL algorithms with sequence models remain domain-specific. 
Therefore, a unified framework capable of encompassing various RL scenarios is an area of future research that requires attention. 
Notably, GPT and multi-modal BeITv3~\cite{BEIT3} have demonstrated a trend toward unifying upstream and downstream tasks and achieved remarkable results.
Although UniMask~\cite{carroll2022unimask} is trying to unify upstream and downstream tasks in RL, it still falls short in performance. 
Thus, the development of a unified modeling approach in sequential decision-making domains will continue to be a critical issue.

In the context of large-scale pre-training, effectively incorporating multi-modal knowledge of vision and language into sequential decision-making is of utmost importance.
While semantic common-sense information plays a critical role in enhancing the efficiency and effectiveness of sequential decision-making for general purposes, ChatGPT~\cite{qin2023chatgpt} achieves a remarkable breakthrough as a powerful knowledge base.
However, the integration between sequential decision-making and perception modalities still lacks naturalness. 
For example, LM-Nav and SayCan manually design the fusion mechanism of multiple outputs from large perception models, but fail to perform joint training. 
While Gato performs joint training of multiple modalities, it lacks alignment between modalities in terms of tasks. 
It would be interesting to explore the possibility of learning an extra module to splice cross-modal large models together, such as an adapter or an inverse dynamic model. 
Furthermore, the emergence of in-context learning and chain-of-thought abilities~\cite{wei2022chain} in large language models may be the ingredients for creating general self-improving agents without the need for supervised knowledge from humans.

\subsection{Efficient Training Systems}
Training Efficiency represents a significant impediment that hinders the development of large decision models, which requires additional efforts and extensive exploration in future research aimed at designing efficient training systems.

\subsubsection{Requirements for Offline Pre-Training}
Although the offline pre-training of decision models shares similarities in data flow and control flow with language and vision transformers, data loading might impede the scalability of the former. 
The offline datasets for large-scale models may exceed the capacity of host memory or even the hard drives of a compute node, necessitating to be served over networks. 
In NLP and CV tasks, training datasets can be shuffled prior to the training phase and sequentially read during the epochs, and thus they can be efficiently cached and accelerated due to underlying space locality. 
However, decision models' performance and stability depend on the data distribution, consequently requiring runtime sampling. 
The random access behavior of sampling might cause a high miss rate for vanilla caching policies and poor performance in data loading. 
Therefore, future researches for training systems with efficient data placement and caching are crucial for pre-training large decision models.

\subsubsection{Requirements for Online Training }
Large decision models impose unique workload characteristics in online training due to the RL paradigm. 
During the online learning phase, since these models periodically interact with the environment and collect mini-batches of training data, these models have to switch repeatedly between inference and training modes. 
As a result, while more researchers and industries have been utilizing high-end GPUs shipped with large GPU memory to accommodate model parameters, their computing resources are often underutilized in distributed training.

Modern GPUs are designed for parallel and batch execution, whereas most existing environments are CPU-oriented and executed sequentially. 
Although the overall environmental throughput can be greatly extended in multi-core systems, a throughput gap may still exist between many CPU-served environments and GPU-served models. 
Therefore, extra abstraction layers and implementation are required to efficiently parallelize and distribute CPU environments. 

Besides, since GPU-served models and CPU-served environments have bi-directional dependencies on each other, their overall performance should be optimized from a systematic view.
For example, larger batches may lead to high peak utilization of devices, while the latency from batching, environment scheduling, and network communication can result in a poor average utilization rate of devices. 
Moreover, frequent communication and synchronization for mode coordination can be expensive in large-scale training. The parameter server, a common component for the asynchronous training paradigm, can easily become a bottleneck for massive parameters and large clusters. 
Therefore, joint efforts in training system design and algorithms are indispensable to address these issues.

\section{Conclusions}\label{sec:summary}
In this survey, we explored the current progress of leveraging the sequence modeling
methods for sequential decision-making tasks.
Tackling sequential decision-making problems via sequence modeling can be a promising solution to address those long-lasting issues in conventional RL methods, involving sample efficiency, credit assignment, and partial observability.
Besides, sequence models can bridge the gap between RL and offline self-supervised learning in terms of data efficiency and transferability.

We conclude that model architecture for large decision models should be designed with the awareness of support for multi-modality, multi-task transferability, and sparse activation, while the algorithms should address the concerns about both the quality and quantity of data.
And the overall training efficiency should be systematically optimized via parallelism.
Following a series of discussions about the theoretical foundation, network architecture, algorithm design and training system support, this survey provides potential research directions toward building a large decision model.
We hope this survey could inspire more investigation into this trending topic and ultimately empower more real-world applications, e.g., robotics, automatic vehicles, and the automated industry.

\begin{acknowledgement}
The SJTU team is partially supported by ``New Generation of AI 2030'' Major Project (2018AAA0100900), Shanghai Municipal Science and Technology Major Project (2021SHZDZX0102) and National Natural Science Foundation of China (62076161). Muning Wen is supported by Wu Wen Jun Honorary Scholarship, AI Institute, Shanghai Jiao Tong University.
\end{acknowledgement}

\bibliographystyle{fcs}
\bibliography{main}

\end{sloppypar}

\end{document}